\documentclass{article}

     \PassOptionsToPackage{numbers,compress}{natbib}

\usepackage[preprint]{neurips_2020}

\usepackage[utf8]{inputenc} 
\usepackage[T1]{fontenc}    
\usepackage[colorlinks=true]{hyperref}       
\usepackage{url}            
\usepackage{booktabs}       
\usepackage{amsfonts}       
\usepackage{nicefrac}       
\usepackage{microtype}      

\usepackage{graphicx}
\usepackage{caption}
\usepackage{subcaption}
\usepackage{xcolor}
\usepackage{stix}

\title{On the Empirical Neural Tangent Kernel of Standard Finite-Width Convolutional Neural Network Architectures}

\author{%
  Maxim Samarin, Volker Roth, David Belius\\
  Department of Mathematics and Computer Science\\
  University of Basel, Switzerland \\
  \texttt{\{\href{mailto:maxim.samarin@unibas.ch}{maxim.samarin}, \href{mailto:volker.roth@unibas.ch}{volker.roth}, \href{mailto:david.belius@unibas.ch}{david.belius}\}@unibas.ch} \\
}

\begin{document}

\maketitle

\begin{abstract}
The Neural Tangent Kernel (NTK) is an important milestone in the ongoing effort to build a theory for deep learning. Its prediction that sufficiently wide neural networks behave as kernel methods, or equivalently as random feature models, has been confirmed empirically for certain wide architectures. It remains an open question how well NTK theory models standard neural network architectures of widths common in practice, trained on complex datasets such as ImageNet. We study this question empirically for two well-known convolutional neural network architectures, namely AlexNet and LeNet, and find that their behavior deviates significantly from their finite-width NTK counterparts. For wider versions of these networks, where the number of channels and widths of fully-connected layers are increased, the deviation decreases.
\end{abstract}

\section{Introduction}

The Neural Tangent Kernel (NTK) is a powerful theoretical tool to model neural networks. To what extent does it model the high performance of the architectures that caused the break-through of deep learning, e.g. AlexNet trained on ImageNet? In the present work we address this question empirically.

Jacot \textit{et al.} \cite{jacot_neural_2018} proved that if one models neural network training as gradient flow (full batch gradient descent of infinitesimal step size), then the training trajectory satisfies an ordinary differential equation (ODE) involving the finite-width Neural Tangent Kernel, a kernel encoded by the architecture of the network and the current time-dependent weights. Furthermore, they showed that if one scales the learning rate per layer in an appropriate way (the "NTK parametrization"), and lets the width tend to infinity this kernel converges to the infinite-width NTK which is \emph{independent} of the weights and \emph{stays constant} during training, greatly simplifying the ODE in this limit. They showed that for $L^2$ loss the predictor at convergence is precisely what a kernel regression using the infinite-width NTK would produce. Importantly, the NTK depends only on the architecture of the network and is not learned. These results were later extended from fully connected architectures to more general ones, including convnets \cite{arora_exact_2019,yang_scaling_2020}. 

Chizat \textit{et al.} \cite{chizat_lazy_2019} pointed out that this result can be understood as the convergence of wide networks to random feature models. Using their point of view, let $f: \mathbb{R}^p \times \mathbb{R}^d \to \mathbb{R}^L$ be the function encoded by a network with weights $w \in \mathbb{R}^p$ and input $x \in \mathbb{R}^d$, and let $f_l$ be the output in component $l=1,\ldots,L$. For $w$ sufficiently close to the random initial weights $w_0$ the first order Taylor expansion in the weights
\begin{equation}\label{eq: taylor}
	f_l(w,x) \approx f_l(w_0, x) + (w-w_0) \cdot \nabla_w f_l(w_0, x), \, l=1,\ldots,L,
\end{equation}
is an accurate approximation. The right-hand side is a random feature model with weights $u = w - w_0$ and the feature embedding of a data point $x$ is given by the gradients $\nabla_w f_l(w_0, x),l=1,\ldots,L,$ with respect to the weights of the original neural network at initialization.
We denote this random feature model by $T_f:\mathbb{R}^p\times\mathbb{R}^d\to\mathbb{R}^L$  with weights $u\in\mathbb{R}^p$. If the approximation (\ref{eq: taylor}) is accurate enough, then also the gradients of the two models $f(w,x)$ (w.r.t. to $w$) and $T_f(u,x)$ (w.r.t. to $u$) will be close. It is therefore intuitive that if one trains $f(w,x)$ with initial weights $w_0$ and $T_f(u,x)$ with initial weights $u=0$ using some form of gradient descent with sufficiently small step size for a sufficiently small number of steps, then the training trajectories will stay close, as long as the weight vector stays in the region around $u=0$ or $w=w_0$, respectively, where the approximation (\ref{eq: taylor}) remains accurate. If the $L^2$ loss is used and the models are overparametrized, one can expect both models to converge to zero loss. If this happens before leaving this region, then the models, whether trained with early stopping or until convergence, will predict a similar function. The NTK result can be proven by showing that for very wide neural networks the models $f(w,x)$ and $T_f(u,x)$ \emph{do} reach zero loss and thus stop evolving before leaving the region where the approximation (\ref{eq: taylor}) is accurate \cite{chizat_lazy_2019}.

As for the infinite-width NTK, the linearized model $T_f$ does not learn a representation but rather uses the representation $\nabla_w f_l(w_0,x),l=1,\ldots,L$ which is fixed by the random initial weights $w_0$ and stays constant throughout training. Thus if training is indeed well-modeled by NTK in a strict sense, then a network and its linearization behave similarly, and no significant feature learning takes place, which is at odds with the usual intuition that neural networks learn good representations. It is thus important to determine how well NTK theory with constant kernel models neural networks of finite-width. In the rest of the paper, "NTK" will always refer to constant kernels given either by finite-width NTK at initialization or the infinite-width NTK, unless stated otherwise. Two extreme scenarios are that (A) very significant feature learning takes place in training standard networks, and the linearized models perform significantly worse or (B) the effect of feature learning in standard neural networks is negligible, and in fact their linearizations perform well and the random feature embedding of $T_f$ is the main reason for good generalization. It is likely that the truth lies somewhere in between, and how close it is to (A) or (B) may depend on the architecture. The goal of our present work is to provide evidence for how close to (A) and (B) some practical convolutional neural network architectures are.

\subsection{Our contributions}

We compare the performance of two standard non-linearized convolutional neural network architectures, LeNet and AlexNet, to the performance of their linearizations $T_f$. We train the networks on MNIST \cite{lenet_98}, CIFAR-10 \cite{CIFAR_2009} and on a subset of ImageNet \cite{ImageNet_2015} with the goal of evaluating how well their performance on these datasets is modeled by NTK theory. We do the same for wider versions of the networks. For LeNet we are able to train networks with a width multiplier of up to $\times$60 and for AlexNet of up to a factor $\times$4. These are the widest networks that would fit in the memory of the GPU we used to train (the number of parameters of a network grows quadratically in the width, and standard width LeNet and AlexNet have about $60k$ and $60m$ parameters, respectively).

Following \cite{lee_wide_2019} we investigate the training trajectories of non-linearized and linearized models by studying the evolution of their output on data points from the validation set. In contrast to the results of \cite{lee_wide_2019} for different architectures, we find that with hyperparameters optimized for validation accuracy of the non-linearized networks, the training trajectories of LeNet immediately diverge, showing the networks are far from the NTK regime. For wider networks the behavior of trajectories of non-linearized and linearized are more similar.

Though LeNet fails this very stringent test of being well-modeled by NTK theory, it is possible that broader characteristics such as final train and test accuracy are in better agreement. To test this we also compare both models trained to convergence. We find that for standard width networks the linearized models has significantly worse test performance. For LeNet trained on MNIST we find a gap of $\sim 6$ percentage points in test performance,  which we consider a large gap since the performance of the linearized network is essentially the same as that of logistic regression on this dataset. For LeNet trained on CIFAR-10 we find a large gap of $\sim 20$ percentage points.
For AlexNet trained on a subset of ImageNet we find a large gap of $\sim 19$ percentage points. The standard-width linearized networks also achieve low train accuracy at $92.9\%$ for LeNet on MNIST, $42.9\%$ for LeNet on CIFAR-10 and $54.7\%$ for linearized AlexNet on a subset of ImageNet. For wider linearized model the gap in both test and train performance closes, but remains large except for LeNet on MNIST. We suspect that the poor training performance of linearized networks is due to the embedding matrix effectively being of low rank, as also argued by \cite{chizat_lazy_2019}, and complement the results of \cite{chizat_lazy_2019} by showing that train accuracy and effective rank (using the measure from \cite{roy_effective_2007}) of the linearized models improves with width.

We conclude that the mechanism behind standard-width AlexNet and LeNet is not well-explained by NTK theory in the strict sense of providing quantitative predictions, since the characteristics of the non-linearized and  linearized networks differ significantly. This supports the view that significant feature learning takes place in typical neural network architectures. The fact that the generalization gap is larger for the more complex datasets CIFAR-10 and ImageNet likely reflects that for these datasets feature learning is more important. Qualitatively, NTK naturally remains a highly insightful analysis of neural network training, and possibly the starting point for theories that do model feature learning \cite{huang_dynamics_2019}.

\subsection{Previous empirical results}

The original NTK paper \cite{jacot_neural_2018} gives experimental results for small synthetic datasets, as well as fully connected networks trained on MNIST of widths $n=100,1000,10^4$, showing good agreement with the infinite-width NTK for the widest network.

Lee \textit{et al.} \cite{lee_wide_2019} carried out experiments for synthetic data, CIFAR-10, and MNIST. For a small datasets (of size $\le256$) they show very good agreement between the linearized model and the original model, with test and train loss, accuracy, and even several individual weights and outputs of the networks at multiple data points tracking each other closely during training (Figures 3 and 4 of \cite{lee_wide_2019}). Similarly good agreement on the same metric was shown for a two hidden layer fully connected network  trained with SGD on full MNIST (Figure S3 \cite{lee_wide_2019}). Most interestingly, a wide ResNet \cite{zagoruyko_wide_2016} trained on the CIFAR-10 shows similar behavior, though the non-linearized model appears to have been trained only to somewhat below $80\%$ training accuracy, and in the test accuracy a gap seems to develop towards the end of training (Figure 7 \cite{lee_wide_2019}).

In \cite{chizat_lazy_2019} a VGG-11 \cite{simonyan_very_2015} widened by a factor $\times6$ and a ResNet-18 \cite{he_deep_2016} widened by a factor $\times7$ were trained on CIFAR-10 with a scaling factor $\alpha$ for tuning the models into the non-linearized and linearized regimes. They observed a large gap in test performance.

Arora \textit{et al.} \cite{arora_exact_2019} devised a method to compute the kernel of the infinite-width NTK of convnet architectures using several standard operations, but not including the max-pooling operation common in standard convnet architectures, and studied the generalization performance of these infinite-width limit NTKs to finite width non-linearized networks. They trained finite-width architectures with two to $20$ convolutional layers and fully-connected (FC) or global average pooling output layers on full CIFAR-10, observing a fairly large gap in test performance.

The experiments most similar to ours, with respect to studying standard convnet architectures, are the Wide ResNet experiments of \cite{lee_wide_2019}, showing good agreement with NTK, at least until the final stage of training, and the widened VGG-11 and ResNet-18 experiments of \cite{chizat_lazy_2019} showing poor agreement. A comparison of results is given in Figure \ref{fig:others-comparison}. Though one should be careful in making a direct comparison due to differences in methodology, it does stand out that the results for widened VGG-11 and ResNet-18 of \cite{chizat_lazy_2019} trained on CIFAR-10 show larger gaps than our less widened AlexNets on a subset of ImageNet, since the latter should be a more complex dataset requiring more feature learning. Determining the source of this discrepancy is beyond the scope of this paper, but possible explanations are differing methodology or that the fact that VGG-11 and ResNet-18 architectures are significantly more efficient at feature learning than AlexNet.

\section{Method}

\begin{figure}[h]
	\centering{}%
	\includegraphics[width=0.78\linewidth]{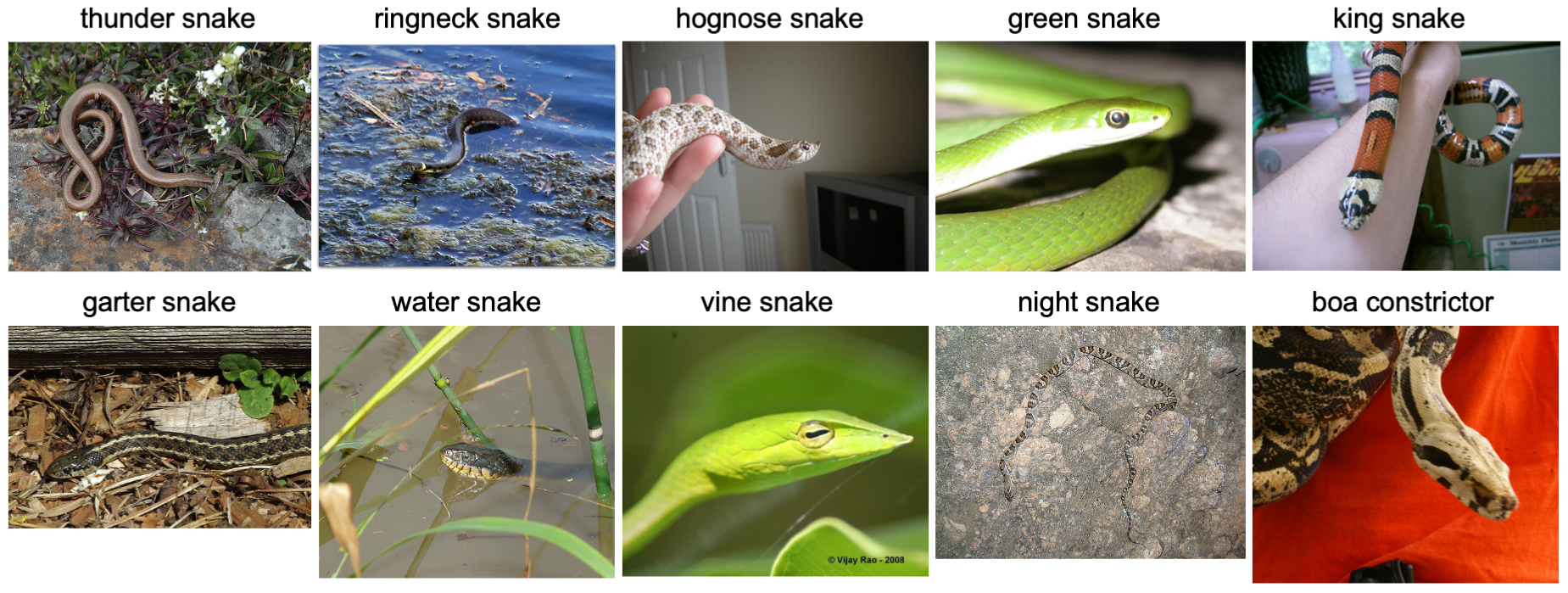}
	
	\begin{tabular}{lcc}
		\toprule 
		\textbf{AlexNet$\mathbf{\times1}$ variations} & \textbf{Train accuracy} & \textbf{Test accuracy} \tabularnewline
		\midrule 
		CE loss & $98.5$ & $51.4$ \tabularnewline 
		$L^2$ loss & $99.1$ & $53.8$ \tabularnewline
		Pre-trained weights & $-$ & $47.6$ \tabularnewline
		\midrule
	\end{tabular}
	\caption{\textbf{Overview of the ten snake categories extracted from ImageNet2012 used for training AlexNet and benchmark performance results}. Training AlexNet on the snakes dataset with $L^2$ loss performs on a par with cross-entropy loss, both exceeding the top-1 test accuracy of pre-trained AlexNet (trained on full ImageNet).}
	\label{fig:snake_overview}
\end{figure}

Our implementation makes use of PyTorch's \cite{pytorch_2019} standard modules for defining and training neural networks.\footnote{Upon acceptance, we intend to make our code publicly available.} For LeNet we adapt the original LeNet-5 architecture of \citep{lenet_98} to use max-pooling and ReLU activations. For AlexNet \cite{alexnet_2012}  we use the PyTorch implementation, which follows \cite{alexnetV2_2014}, with $10$ outputs rather than $1000$ (see below). Despite training for classification we use the $L^2$ loss with one-hot encoded target vectors. Firstly, with standard cross-entropy loss the networks never converge to exactly zero loss, so the networks must at some point leave the region where the approximation (\ref{eq: taylor}) is valid, causing some ambiguity in the heuristic. Secondly, $L^2$ loss allows for an easier and more efficient implementation of the training of the linearized models. We furthermore disable dropout for AlexNet, since it not clear to us how to model it in the NTK framework (see however \cite{novak_neural_2020}). We find that that after optimizing hyperparameters we can train LeNet and AlexNet as described above to train and test performance similar to their performance when trained with cross-entropy loss without dropout (see Figures \ref{fig:snake_overview} and \ref{fig:lenet-mnist-width}). We predict the class whose one-hot vector is closest to the output vector, which is equivalent to predicting the argmax of the output layer.

\begin{table}[h]
	\centering
	\begin{tabular}{llcccc}
		\toprule 
		\textbf{Architecture} & \textbf{Dataset} & \textbf{Standard} & \textbf{Linearized} & \textbf{Generalization gap} \tabularnewline
		\midrule 
		LeNet & MNIST	& $99.2$ & $93.2$ & \hspace{1.5mm}$6.0$ \tabularnewline
		LeNet$\times60$ & MNIST & $99.4$ & $98.9$	& \hspace{1.5mm}$0.5$ \tabularnewline
		LeNet &  CIFAR-10	& $62.5$	& $42.3$	& $20.2$  \tabularnewline
		LeNet$\times60$ & CIFAR-10	& $78.8$ & $65.6$	& $13.2$   \tabularnewline
		VGG-11$\times6$ \cite{chizat_lazy_2019} & CIFAR-10	& $89.7$ & \hspace{1.5mm}$61.7^{*}$ & $28.0$ \tabularnewline
		ResNet-18$\times8$ \cite{chizat_lazy_2019}  & CIFAR-10 & $91.0$ & \hspace{1.5mm}$56.7^{*}$ & $34.3$  \tabularnewline
		21 layer CNN-V \cite{arora_exact_2019} & CIFAR-10	& $75.6$ & \hspace{1.5mm}$64.1^{\dagger}$ & $11.5$ \tabularnewline
		21 layer CNN-GAP \cite{arora_exact_2019} & CIFAR-10	& $83.3$ & \hspace{1.5mm}$77.1^{\dagger}$ & \hspace{1.5mm}$6.2$ \tabularnewline
		AlexNet & ImageNet (subset) & $53.8$ & $35.2$	& $18.6$ \tabularnewline
		AlexNet$\times4$ & ImageNet (subset) & $57.0$ & $39.2$ & $17.8$\tabularnewline
		\bottomrule
	\end{tabular}
	\vspace{3mm}
	\caption{\textbf{Our and previous empirical results}. The stated numbers for standard and linearized are test accuracy results in percent of the different networks, with $\times$\textit{factor} specifying the factor by which the networks were widened. Note that numbers may not be directly comparable due to differing methodologies. $^*$: training using a scaling factor \cite{chizat_lazy_2019}. $^\dagger$: results for the infinite-width NTK.}\label{fig:others-comparison}
\end{table}

In addition, we train linearized versions of LeNet and AlexNet, i.e. the random feature models
\begin{equation}
\big(T_f(u,x)\big)_l = f_l(w_0, x) + u \cdot \nabla_w f_l(w_0, x), \, l=1,\ldots,L,
\end{equation}
where $f(w,x)$ is the function of weights and inputs represented by the original network. We train them with SGD in the standard way by optimizing $u$ with gradient updates coming from
\begin{equation}
\nabla_u |T_f(u, x)-y|^2 = 2 \sum_{l=1}^L \nabla_w f_l(w_0, x) \left( \left(T_f(u,x)\right)_l-y_l\right).
\end{equation}
For brevity we refer to the linearized networks by prepending "Lin", i.e. LinLeNet and LinAlexNet. Finally, we train wider versions of the above models -- both linearized and non-linearized -- in which the number of units in each fully-connected layer, and the number of channels in each convolutional layer, are multiplied by widening factor denoted by $\times$\textit{factor}.  

We train LeNet on the full MNIST and CIFAR-10 datasets. Computing the the gradients of the linear model with $L$ outputs requires computing $L$ gradients of the original network per data point, and thus $L$ backward passes, which is computationally intensive if $L$ is large. We therefore train AlexNet and LinAlexNet on a subset of ImageNet consisting of $L=10$ classes.

As our goal is to stay as close as possible to standard neural network training practices, we use SGD with weight decay and momentum and the standard PyTorch weight initialization, which is a variant of Kaiming initialization \cite{he_delving_2015}, rather than the NTK parametrization that is used in the proof of \cite{jacot_neural_2018}. We do hyperparameter search individually for each non-linearized network and each widening factor.

\section{Experiments}

We train on MNIST, CIFAR-10, and a subset of ImageNet which contains ten different snake classes. We deliberately choose similar classes to form a challenging classification task. The extracted dataset contains 1300 training and 50 test images per class, resulting in 13000 train and 500 test images in total. Figure \ref{fig:snake_overview} provides an overview of the dataset and benchmark performance results indicating that training with $L^2$ loss performs on par with cross-entropy loss (in both settings no dropout is used). Each result represents a single run of the specified network for 100 epochs unless stated otherwise. Most computations were conducted on Nvidia GeForce Titan X Pascal and Tesla V100 GPUs. For experiments involving LinAlexNet$\times3$ and LinAlexNet$\times4$ we used an Nvidia Quadro RTX 8000 with 48 GB memory due to the increased memory requirement.

\subsection{Early training trajectory experiments for different widths}
\begin{figure}[h]
    \centering
    \includegraphics[width=\textwidth]{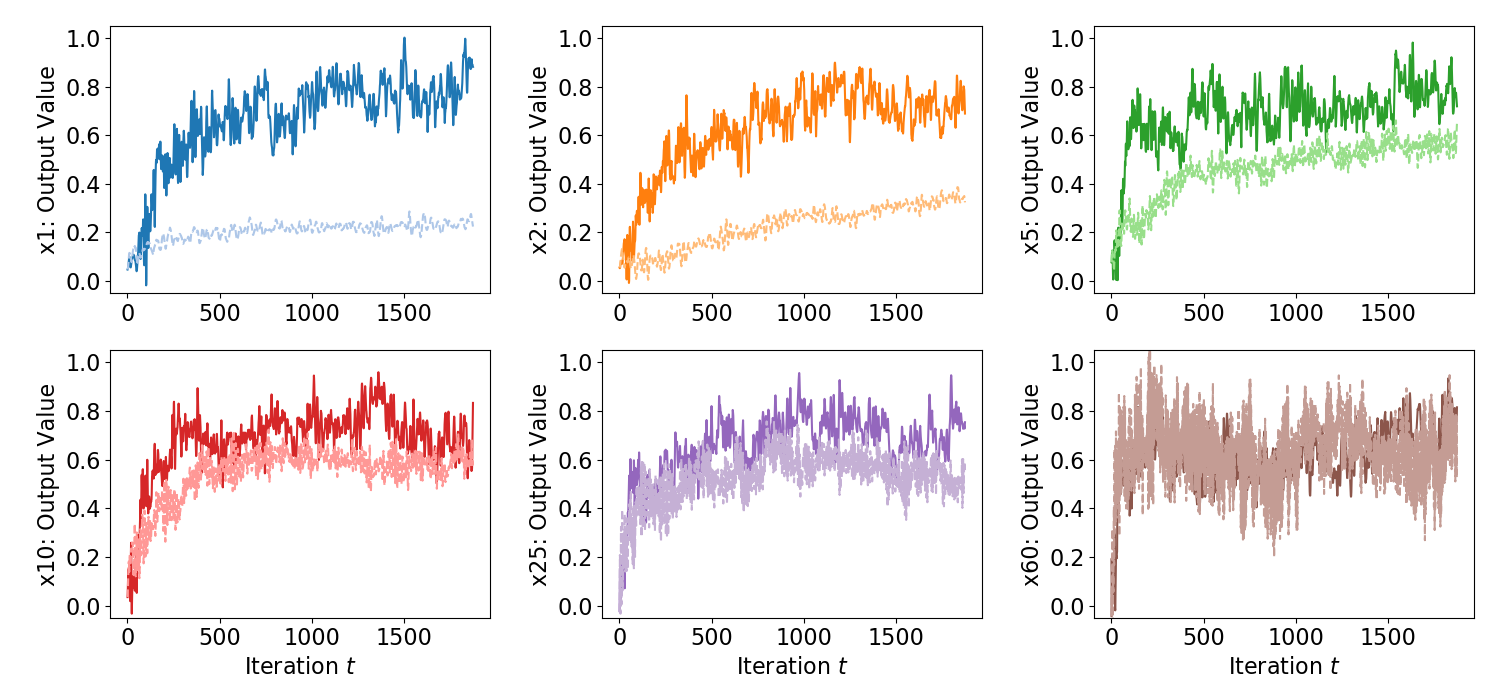}
    \caption{\textbf{Training trajectories of LeNet and LinLeNet do not stay close for small width factors}. The trajectory of output values for the same example from the validation set for different widths of (Lin)LeNet trained on MNIST. Light colors indicate the LinLeNet output values.}
    \label{fig:LeNet_output_trajectories}
\end{figure}

In this section we report on experiments tracking the output of the linearized and non-linearized networks throughout training, to observe if they stay close. Following \cite{lee_wide_2019}, we pick several data points $x$ of the validation set and plot $f_l(w_t, x)$ and  $\left( T_f(u_t, x) \right)_l$ against $t=0,1,\ldots$, where $w_t$ and $u_t$ are the weights after $t$ gradient updates for LeNet and LinLeNet on MNIST. The output $l$ is the output for the correct class of the data point $l$. We use batch size $32$, momentum $0.9$, weight decay $5\times 10^{-5}$, learning rate $0.1$ (more on the hyperparameter selection below), and a fixed random seed to ensure that both networks with a particular width factor start from the same initialization and receive the same minibatches. The results are shown in \ref{fig:LeNet_output_trajectories}. We see that the trajectories immediately diverge at (standard) width factor $\times1$. For larger width factors the curves behave increasingly similarly.

Certainly for width factors $\times1$, $\times3$, and $\times 5$ our results show that LeNet is not well-modeled by NTK theory in the strict sense of training trajectories of non-linearized and linearized models staying close.

\subsection{Train and generalization error at convergence for LeNet and LinLeNet}

In this section we present experiments comparing standard LeNet and its random feature model LinLeNet with width multipliers ranging from $1$ to $60$ on MNIST and CIFAR-10. For each width of LeNet we train for $100$ epochs with SGD using batch size $32$, weight decay $5\times10^{-5}$, momentum $0.9$, and learning rates in $\{0.1, 0.01, 0.001\}$. For CIFAR-10 we in addition decreased the learning rate by a factor $10$ every 30 epochs. For all widths and both datasets a learning rate of $0.1$ had highest test accuracy, except for MNIST with width factors $\times2$, $\times5$, and $\times10$ where $0.01$ performed better by at most $0.08$ percentage points, likely within range of random fluctuation. In addition, LinLeNet was trained with learning rate $1$. For LinLeNet on MNIST learning rate $0.1$ performed the best, while for LinLeNet on CIFAR-10 at certain width factors 
a learning rate of $1$ performed better than $0.1$ by a small margin: for $\times2$ and $\times5$ by 0.1 percentage points, and for $\times1$ and $\times10$ by 0.2 percentage points. For some  widths of LeNet on both datasets we also tried a learning rate of $1$, for which training did not converge or test performance was worse. All the presented results are for learning rate $0.1$.

\begin{figure}[h]
	\begin{centering}
	\includegraphics[width=0.45\linewidth]{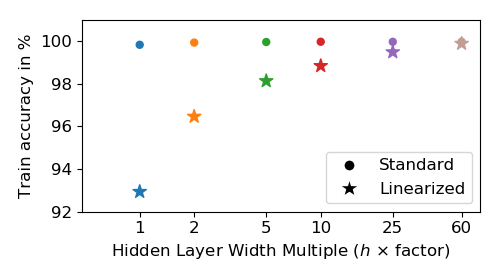}
	\includegraphics[width=0.45\linewidth]{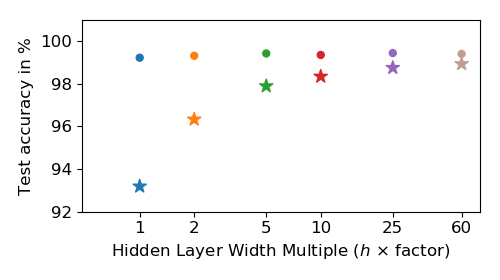}
	\par\end{centering}
	\centering{}%
	\vspace{3mm}
	\begin{tabular}{llllllll}
		\toprule 
		\textbf{Architecture} & \textbf{Set} & $\mathbf{\times1}$ & $\mathbf{\times2}$ & $\mathbf{\times5}$ & $\mathbf{\times10}$ & $\mathbf{\times25}$ & $\mathbf{\times60}$ \tabularnewline
		\midrule 
		LinLeNet & Test & $93.2$ & $96.3$ & $97.9$ & $98.3$ & $98.8$ & $98.9$ \tabularnewline
		LeNet & Test & $99.2$ & $99.3$ & $99.4$ & $99.4$ & $99.4$ & $99.4$ \tabularnewline
		\midrule 
		LinLeNet & Train & $92.9$ &$96.5$ &$98.1$ &$98.8$ &$99.5$ &$99.9$  \tabularnewline
		LeNet & Train & $99.8$ & $99.9$ & $100$ & $100$ & $100$ & $100$ \tabularnewline
		\bottomrule
	\end{tabular}
	\caption{The train and test accuracy of LeNet ($\bullet$) and LinLeNet ($\star$) trained on MNIST for different widths.}
	\label{fig:lenet-mnist-width}
\end{figure}

\begin{figure}
	\begin{centering}
	\includegraphics[width=0.47\linewidth]{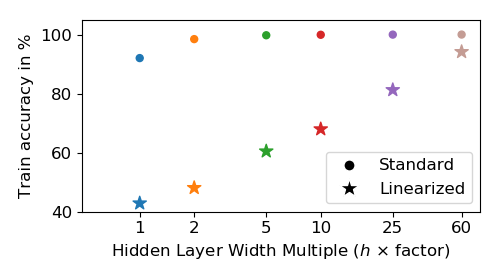}
	\includegraphics[width=0.47\linewidth]{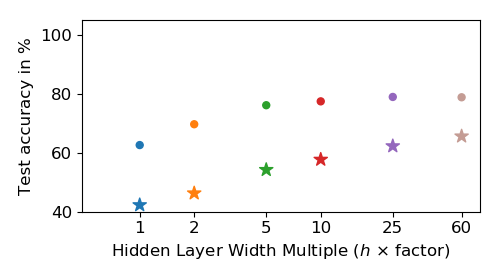}
	\par\end{centering}
	\centering{}%
	\vspace{3mm}
	\begin{tabular}{llllllll}
		\toprule 
		\textbf{Architecture} & \textbf{Set} & $\mathbf{\times1}$ & $\mathbf{\times2}$ & $\mathbf{\times5}$ & $\mathbf{\times10}$ & $\mathbf{\times25}$ & $\mathbf{\times60}$ \tabularnewline
		\midrule 
		LinLeNet & Test & $42.3$ & $46.3$ & $54.2$ & $57.7$ & $62.2$ & $65.6$ \tabularnewline
		LeNet & Test & $62.5$ & $69.6$ & $76.1$ & $77.4$ & $78.9$ & $78.8$ \tabularnewline
		\midrule 
		LinLeNet & Train & $42.9$ &$48.1$ &$60.5$ &$68.0$ &$81.2$ & $94.1$  \tabularnewline
		LeNet & Train & $92.0$ & $98.4$ & $99.8$ & $99.9$ & $100$ & $100$ \tabularnewline
		\bottomrule
	\end{tabular}
	\caption{The train and test accuracy of LeNet ($\bullet$) and LinLeNet ($\star$) trained on CIFAR-10 for different widths.}
	\label{fig:lenet-cifar-width}
\end{figure}

The results for MNIST are presented in Figure \ref{fig:lenet-mnist-width}.
For the standard width a substantial difference of $\sim 6.5$ percentage points in generalization error between LeNet and LinLeNet is observed. While LeNet does not gain appreciably from increasing the width, LinLeNet does, and the gap shrinks to $\sim 0.5$ percentage points for width factor $\times60$. For factor $\times1$ the linearized model performs essentially like logistic regression on normalized MNIST pixels, which achieves about $93\%$ train and $92\%$ test accuracy.

Results for CIFAR-10 are presented in Figure \ref{fig:lenet-cifar-width}. For the standard width a difference of $\sim 20$ percentage points in generalization error between LeNet and LinLeNet is observed. This shrinks to a smaller but still significant gap of $\sim 13$ percentage points at width factor $\times60$. Standard width (non-linearized) LeNet achieves only $92\%$ train accuracy, suggesting that it is underparameterized for CIFAR-10.

\begin{figure}[h]
	\begin{centering}
	\includegraphics[width=0.43\linewidth]{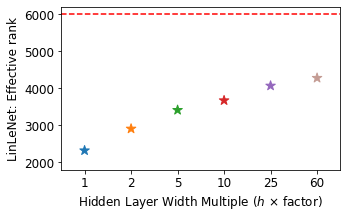}
	\includegraphics[width=0.43\linewidth]{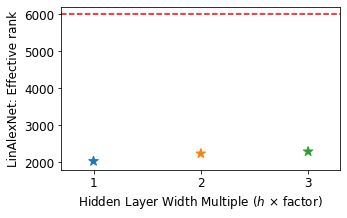}
	\par\end{centering}
	\caption{Effective rank of the matrix of feature embeddings for $600$ data examples for LinLeNet on MNIST and LinAlexNet on our ImageNet subset. The matrix has $6000$ rows.}
	\label{fig:eff_rank_lenet_mnist}
\end{figure}

In both experiments qualitatively similar gaps are observed for the train accuracy, which raises the question if the linearized models are well-trained. Note, however, that fitting the linearized model with $m$ data points is effectively solving a linear system $Au = y$ for weights $u\in\mathbb{R}^p$ with target $y \in \mathbb{R}^{n}$, where $p$ are the number of parameters of the original model and the rows of the matrix $A$ are gradients of each output of the network at each data point. Matrix $A$ has $n=10 \times m$ rows  since one must fit each of the ten outputs for each data point. Thus, for LinLeNet at width factors $\times1$ and $\times2$ with roughly $60k$ and $240k$ parameters, respectively, matrix $A$ can not have full rank when fitting a dataset of size $m=50k$, making it impossible to fit arbitrary targets. Moreover, even for wider networks it appears that $A$ remains effectively of low rank. We show this by computing the effective rank \cite{roy_effective_2007} of the matrix for $600$ data points and each width factor. These effective ranks are much lower than the number of rows which is $6000$, and increases with width factor as is illustrated in Figure \ref{fig:eff_rank_lenet_mnist}. We suspect that to interpolate the training data in MNIST and CIFAR-10 with LinLeNet one needs to fit $u$ also in a subspace with very small singular values, making it difficult to achieve close to $100\%$ train accuracy with SGD.

The large gaps observed for linearized and standard (non-linearized) LeNet$\times1$  suggests that at the standard width significant feature learning is taking place in the standard (non-linearized) model. The closing of the gap suggests that at width factor $\times60$ feature learning is not very significant for MNIST, but remains significant for the more complex CIFAR-10 dataset, which is consistent with the intuitive view that more complex datasets require more significant feature learning.

\subsection{Train and generalization error at convergence for AlexNet and LinAlexNet}

In this section we present experiments comparing non-linearized AlexNet and its linearized version LinAlexNet for width factors $\times1, \times2, \times3$, and $\times4$ trained on the 10-class snakes subset of ImageNet2012 \cite{ImageNet_2015} (see Figure \ref{fig:snake_overview}).

For the non-linearized networks we used batch size $32$, weight decay $5\times10^{-6}$, momentum $0.9$, learning rates in $\{1, 0.1, 0.01, 0.001\}$ and trained for $100$ epochs, decreasing the learning rate by a factor $10$ every 30 epochs. For all widths the learning rate $0.1$ gave the best test performance.

Trained with the same hyperparameters as their non-linearized counterparts, which would be appropriate if the model training trajectories stay close together, the gap in generalization error between standard (non-linearized) AlexNet$\times1$ and linearized AlexNet$\times1$ is very large at $\sim 25$ percentage points. We find that increasing the width of AlexNet leads to a small increase in test performance, while it leads to a greater increase in the generalization performance of LinAlexNet. At width factor $\times4$ the gap is $\sim 21$ percentage points.

\begin{figure}
	\begin{centering}
	\includegraphics[width=0.45\linewidth]{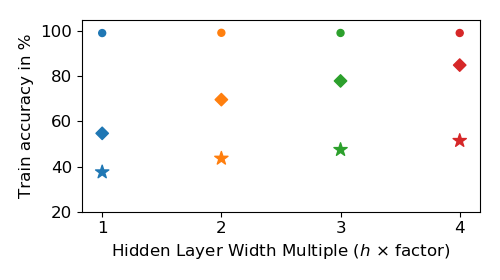}
	\includegraphics[width=0.45\linewidth]{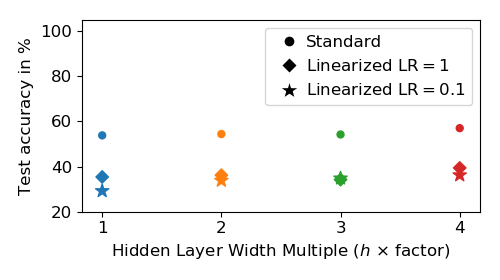}
	\par\end{centering}
	\centering{}%
	
	\begin{tabular}{llllll}
		\toprule 
		\textbf{Architecture} & \textbf{Set} & $\mathbf{\times1}$ & $\mathbf{\times2}$ & $\mathbf{\times3}$ & $\mathbf{\times4}$ \tabularnewline
		\midrule 
		LinAlexNet LR $1$ & Test & $35.2$ & $36.4$ & $34.2$ & $39.2$ \tabularnewline
		LinAlexNet LR $0.1$ & Test & $29.2$ & $33.8$ & $34.8$ & $36.4$ \tabularnewline
		AlexNet & Test & $53.8$ & $54.4$ & $54.2$ & $57.0$  \tabularnewline
		\midrule 
		LinAlexNet LR $1$ & Train & $54.7$ & $69.6$ & $77.9$ & $84.9$   \tabularnewline
		LinAlexNet LR $0.1$ & Train & $37.5$ & $43.6$ & $47.5$ & $51.5$   \tabularnewline
		AlexNet & Train & $99.1$ & $99.2$ & $99.2$ & $99.2$  \tabularnewline
		\bottomrule
	\end{tabular}
	
	\caption{The train and test accuracy of AlexNet ($\bullet$), LinAlexNet with learning rate $0.1$ ($\star$), and LinAlexNet with learning rate $1$ ({\scriptsize$\mdblkdiamond$}) trained on the snakes dataset for different widths.}\label{fig:alexnet-snakes-width}
	\vspace{-3mm}
\end{figure}

We also trained the linearized models with a larger learning rate of $1$ and weight decay in $\{5\times 10^{-6}, 5\times 10^{-7}, 5\times 10^{-8}\}$, and found that this improved test and train accuracy. The results for weight decay $5\times 10^{-7}$ (which gave the best test performance) are shown in Figure \ref{fig:alexnet-snakes-width}. The generalization gap with the non-linearized models shrinks to $\sim 18$ percentage points both for the standard width $\times1$ and for width factor $\times4$.

For the linearized models the train accuracies are very low at all widths. Thus these models are far from interpolating on the training set, despite having many more parameters than the number of equations in the system $Au=y$ they aim to solve, in which $A$ has $130k$ rows and $u$ has dimension at least $60m$. As described above for LeNet we believe that this is due to $A$ being of low rank, effectively. The effective rank \cite{roy_effective_2007} of the matrix $A$ for $600$ examples is shown in Figure \ref{fig:eff_rank_lenet_mnist} (right) where at all widths it is significantly lower than the number of rows of $A$ which is $6000$.

The significantly different training and test accuracies, as well as different optimal hyperparameters, shows that at all widths considered AlexNet is far from being in the NTK regime.

\section{Conclusion}

We applied established training procedures to standard AlexNet trained on a subset of ImageNet and to standard LeNet trained on MNIST as well as CIFAR-10 and find that these training procedures are not well-modeled by NTK theory with a constant kernel. Training trajectories do not stay close even for LeNet on MNIST, and there are significant gaps in train and test performance between non-linearized and linearized models. This remains true for wider versions of these networks, except for LeNet on MNIST with large width factors.

It is possible that with a much smaller learning rate which better approximates the gradient flow of theoretical results, the agreement would be better. While we cannot rule this out based on our experiments, we believe the main reason for the poor agreement is that significant feature learning takes place in training these network architectures, which is not modeled by NTK theory with constant kernel. The only exception we forsee is the generalization performance of infinite-width LeNet NTK on MNIST, which could well be very close to that of finite-width non-linearized LeNet.

Despite our results showing that NTK theory cannot explain the performance of standard AlexNet and LeNet, determining the performance of infinite-width AlexNet and LeNet kernels on various datasets remains a very interesting open question. It is possible that at infinite-width the performance gap closes almost completely, though we find this scenario unlikely in the case of AlexNet.

First and foremost our results highlight the need for a theory that goes beyond NTK with constant kernel, for instance by modeling the change of the time-dependent NTK \cite{huang_dynamics_2019,jacot_neural_2018} for finite-width networks, or by further developing the various proposed "mean-field" theories \cite{chizat_global_2018,hu_mean-field_2019,javanmard_analysis_2019,mei_mean-field_2019,nguyen_mean_2019,rotskoff_global_2019}.

\section{Acknowledgement}

We are grateful to Levent Sagun and Ivan Dokmani\'{c} for enlightening discussions regarding the results presented in this article. M.S. would like to thank the Swiss National Science Foundation for supporting the research with grant 407540 167333 as part of the Swiss National Research Programme NRP 75 "Big Data". Calculations were performed at sciCORE (http://scicore.unibas.ch/) scientific computing core facility at University of Basel and Amazon Web Services (AWS).

\bibliography{bib}

\end{document}